%% file: main.tex
\documentclass[10pt,twocolumn,letterpaper]{article}

\usepackage[final]{cvpr}      

\usepackage{graphicx}
\usepackage{amsmath}
\usepackage{amssymb}
\usepackage{booktabs}
\usepackage{multirow}

\usepackage{algorithm}
\usepackage{algpseudocode} 

\usepackage[normalem]{ulem}
\usepackage{bm}
\usepackage{wrapfig}
\usepackage[table]{xcolor}

\usepackage{enumitem} 

\usepackage[pagebackref,breaklinks,colorlinks]{hyperref}

\usepackage[capitalize]{cleveref}
\crefname{section}{Sec.}{Secs.}
\Crefname{section}{Section}{Sections}
\Crefname{table}{Table}{Tables}
\crefname{table}{Tab.}{Tabs.}

\begin{document}

\NewDocumentCommand\tomato{}{{\includegraphics[scale=0.025]{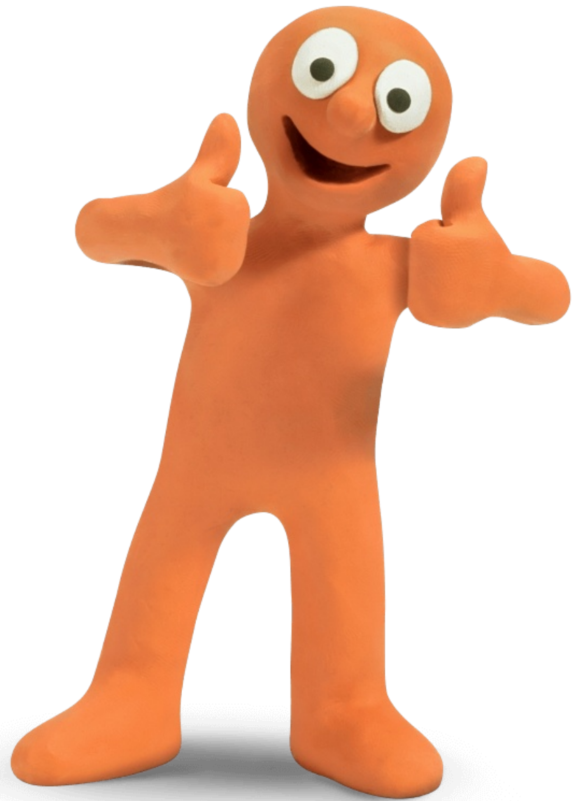}}}
\newcommand{\todo}[1]{\textcolor{red}{[\textbf{TODO:} #1]}}
\newtheorem{example}{Example}
\title{\tomato Morph: A Motion-free Physics Optimization Framework \\ for Human Motion Generation \vspace{-1em}}

\author{%
  \small {Zhuo Li}\thanks{Equal contribution}~~$^{1}$, {~Mingshuang Luo}$^{*~2,3,4}$, \,{~Ruibing Hou}\thanks{Corresponding author}~~$^{2}$, {~Xin Zhao}$^{5}$, \\ \small {~Hao Liu}$^{1}$, {~Hong Chang}$^{2,4}$, {~Zimo Liu}$^{3}$, {~Chen Li}$^{1}$\\ 
  \small {$^1$WeChat, Tencent Inc},
  \small {$^2$Key Laboratory of Intelligent Information Processing of Chinese Academy of Sciences (CAS),} \\ \small {Institute of Computing Technology, CAS, China} \\
  \small {$^3$Peng Cheng Laboratory, China}, \small {$^4$University of Chinese Academy of Sciences, China} \\  \small {$^5$MoE Key Laboratory of Artificial Intelligence, AI Institute, Shanghai Jiao Tong University} \\
}

\maketitle

\input{sections/0_abstract}
\input{sections/1_introduction}

\input{sections/2_related_work}

\input{sections/3_method}
\input{sections/4_experiment}
\input{sections/5_conclusion}

\small
\bibliographystyle{ieee_fullname}
\bibliography{main}

\newpage
\input{sections/supplementary_material}

\end{document}

%% file: sections/0_abstract.tex
\begin{abstract}

Human motion generation has been widely studied due to its crucial role in areas such as digital humans and humanoid robot control. However, many current motion generation approaches disregard physics constraints, frequently resulting in physically implausible motions with pronounced artifacts such as floating and foot sliding. Meanwhile, training an effective motion physics optimizer with noisy motion data remains largely unexplored.
In this paper, we propose \textbf{Morph}, a \textbf{Mo}tion-F\textbf{r}ee \textbf{ph}ysics optimization framework, consisting of a Motion Generator and a Motion Physics Refinement module, for enhancing physical plausibility without relying on expensive real-world motion data. Specifically, the motion generator is responsible for providing large-scale synthetic, noisy motion data, while the motion physics refinement module utilizes these synthetic data to learn a motion imitator within a physics simulator, enforcing physical constraints to project the noisy motions into a physically-plausible space. Additionally, we introduce a prior reward module to enhance the stability of the physics optimization process and generate smoother and more stable motions. These physically refined motions are then used to fine-tune the motion generator, further enhancing its capability. This collaborative training paradigm enables mutual enhancement between the motion generator and the motion physics refinement module, significantly improving practicality and robustness in real-world applications.
Experiments on both text-to-motion and music-to-dance generation tasks demonstrate that our framework achieves state-of-the-art motion quality while improving physical plausibility drastically. Project page: \url{https://interestingzhuo.github.io/Morph-Page/}. 

\end{abstract}

%% file: sections/1_introduction.tex
 \begin{figure}[!t]
	\centering
	\tabcolsep=0.05cm
    \includegraphics[width=0.99\linewidth]{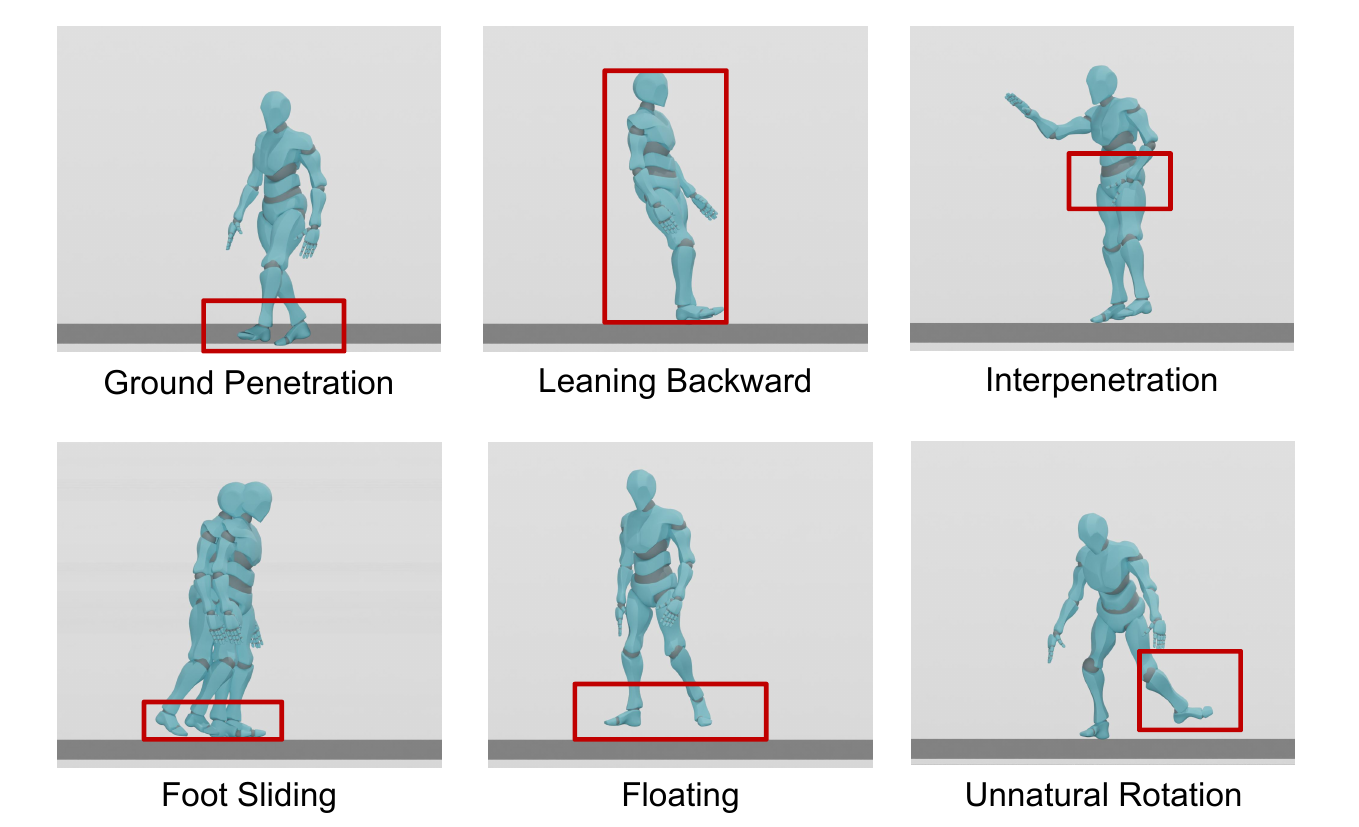}
    \caption{Examples of physical inconsistencies in generations.} 
    \vspace{-0.5cm}
\label{fig:physic_plaus}
\end{figure}

\vspace{-0.2cm}
\section{Introduction}
Accurate human motion synthesis plays a crucial role in applications such as robotics, video gaming, and virtual reality \cite{guzov2021human,von2018recovering,zhang2020generating,zhang2020place,bhatnagar2022behave}. Recent advances in AI have enabled motion generation under various control conditions, including textual descriptions and music inputs. Current approaches can be categorized into three main paradigms: conditional diffusion models \cite{tevet2023human,zhang2023remodiffuse,Yuan_2023_physdiff,tseng2023edge,li2024lodge}, conditional autoregressive models \cite{pinyoanuntapong2024bamm,jiang2024motiongpt,zhang2024motiongpt,zhong2023attt2m,luo2024m3gpt}, and generative masked modeling \cite{guo2024momask,pinyoanuntapong2024mmm}. These models have significantly improved the quality of generated motion by better capturing the complex multimodal distribution of human movement.

However, most existing motion generation approaches overlook a fundamental aspect of human motion: the \textit{laws of physical}. Although these generation models excel in capturing statistical distribution of human motions, they lack explicit mechanisms to enforce physical constrains. As a result, the generated motions frequently exhibit pronounced artifacts such as ground penetration, leaning backward, interpenetration, foot sliding, floating and unnatural rotation, as shown in Fig.~\ref{fig:physic_plaus}. Given human sensitivity to even slight physical inconsistencies, these  physically implausible motions hinder  many real-world applications such as animation and virtual reality \cite{Yuan_2023_physdiff, reitsma2003perceptual, hoyet2012push}. 

Recently, a few studies \cite{Yuan_2023_physdiff, han2024reindiffuse} have attempted to enhance the physical plausibility of motion generation, making the generated motions more realistic. For example, PhysDiff \cite{Yuan_2023_physdiff} and Reindiffuse \cite{han2024reindiffuse} incorporate physical constraints into the denoising diffusion process. However, these approaches face two main limitations. On the one hand, although these works \cite{Yuan_2023_physdiff, han2024reindiffuse} introduce promising approaches for motion physics optimization, their generalizability and transferability across different types of motion generation models and tasks have not been fully demonstrated or explored. On the other hand, these works apply physics and diffusion iteratively, embedding physics optimization into multiple steps of diffusion process. The frequent execution of physics optimization significantly increases computational costs during inference. Additionally, prior studies \cite{luo2021dynamics, Luo2023PerpetualHC, luo2023universal} have explored physics-based human motion imitation using a physics simulator, typically relying on large-scale real 3D motion data as priors to enhance accuracy and naturalness in imitation learning. However, collecting large-scale realistic motion data is challenging. A natural question arises: \textit{Is it possible to learn an efficient, model-agnostic physical optimizer without relying on real motion data?}

To achieve this, we propose a \textbf{Mo}tion-f\textbf{r}ee \textbf{ph}ysics optimization framework, namely \textbf{Morph}. As shown in Fig.~\ref{fig:morph_whole_framework}, Morph consists of two main modules: a Motion Generator \textbf{(MG)} that can be any existing motion generator, and a Motion Physic Refinement \textbf{(MPR)} module for enhancing physical plausibility. Morph employs a two-stage optimization process. \textbf{\textit{In the first stage}}, using large-scale noisy motion data produced by the motion generator, MPR module is optimized to project input motions into  a physically-plausible space. Specifically, MPR module designs a motion imitator that controls a character agent to mimic the given noisy motions within a physics simulator. The simulator enforces multiple physical constraints, effectively reducing artifacts such as floating and foot sliding. To further ensure the naturalness of simulated motions, MPR module introduces a prior reward module to align the distribution of physics-refined motions with that of input motions. This prior reward module can provide smoother and more stable rewards. Feedback signals from both physics simulator and prior reward module guide the optimization of the motion imitator via reinforcement learning.
\textbf{\textit{In the second stage}}, we address the limitations of pretrained motion generators, which may not be trained on data that adheres to physical constraints and thus produce suboptimal results. To this end, we finetune the motion generator using the physic-refined motion data produced by MPR module, further enhancing its generation capability.  
During inference, the fine-tuned motion generator and MPR module work in tandem to generate  physics-plausible and high-quality motions. 

By fostering the synergy between motion generation and motion physics refinement, morph provides a scalable, efficient, and generalizable solution to realistic human motion synthesis. We evaluate our morph framework on text-to-motion and music-to-dance tasks, using diffusion-based, autoregressive, and generative mask models. Extensive experiments demonstrate that morph achieves significant reduces physical errors, while maintaining competitive motion quality across different generators and tasks, despite not being trained on real motion data.

Our contributions can be summarized as follows:
\vspace{-0.2cm}
\begin{itemize}
\item We introduce Morph, a generalizable framework that enhances physical plausibility and generation quality across various motion generation models and tasks. Also, a prior reward module is designed to accelerate imitation learning and improve motion realism in physics simulation.
\vspace{-0.2cm}
\item We design a collaborative training paradigm, where the Motion Generator provides large-scale synthetic motion data to enhance the Motion Physics Refinement module, which, in turn,  refines motion quality by feeding back physics-refined data.
\vspace{-0.2cm}
\item Extensive qualitative and quantitative experiments demonstrate that Morph achieves competitive physical plausibility and generation performance.
\end{itemize}

%% file: sections/2_related_work.tex
\section{Related Work}

\noindent
\textbf{Human Motion Generation.} \
Motion generation is a long-history  task that can be conditioned on various signals, such as  text description, music and action \cite{zhu2023human, guo2024momask, zhang2023t2m, guo2022generating, siyao2022bailando, tseng2023edge, yi2023generating, li2025decoupled, li2025beyond, cen2024generating, dai2024interfusion, tanaka2023role, zhang2024large}. Our work specifically 
 focuses on text-to-motion \cite{guo2022generating, zhang2023t2m, zhang2024motiondiffuse, tevet2023human, guo2024momask, pinyoanuntapong2024mmm} and music-to-dance generation  \cite{tseng2023edge, siyao2022bailando, li2024lodge, marchellus2023m2c}. Mainstream approaches can be roughly divided into three categories: diffusion-based methods \cite{tevet2023human, chen2023executing, zhang2023remodiffuse, Yuan_2023_physdiff, tseng2023edge, li2024lodge}, autoregressive models \cite{zhang2023t2m, luhumantomato, jiang2024motiongpt, zhang2024motiongpt, zhong2023attt2m, luo2024m3gpt, siyao2022bailando} and generative masked modeling  \cite{guo2024momask, pinyoanuntapong2024mmm}. For instance, MDM \cite{tevet2023human} uses a transformer-based diffusion model with conditional text representations extracted from CLIP \cite{radford2021learning}. T2M-GPT \cite{zhang2023t2m} designs a conditional autoregressive transformer model based on VQ-VAE \cite{van2017neural} and GPT \cite{radford2018improving, vaswani2017attention}. MoMask \cite{guo2024momask} employs residual vector quantization and generative masked transformers to iteratively generate motions.  For music-to-dance generation, Bailando \cite{siyao2022bailando} predicts discrete token sequences conditioned on music and uses an autoregressive transformer to regenerate the dance sequence. 
However, existing motion generation models often produce physically implausible motions, as they overlook physical laws during training. In contrast, our proposed MPR module effectively enforces physical constraints, significantly  enhancing the physically plausibility of generated motions.

\begin{figure*}[t]
	\centering
	\tabcolsep=0.05cm
    \includegraphics[width=0.84\linewidth]{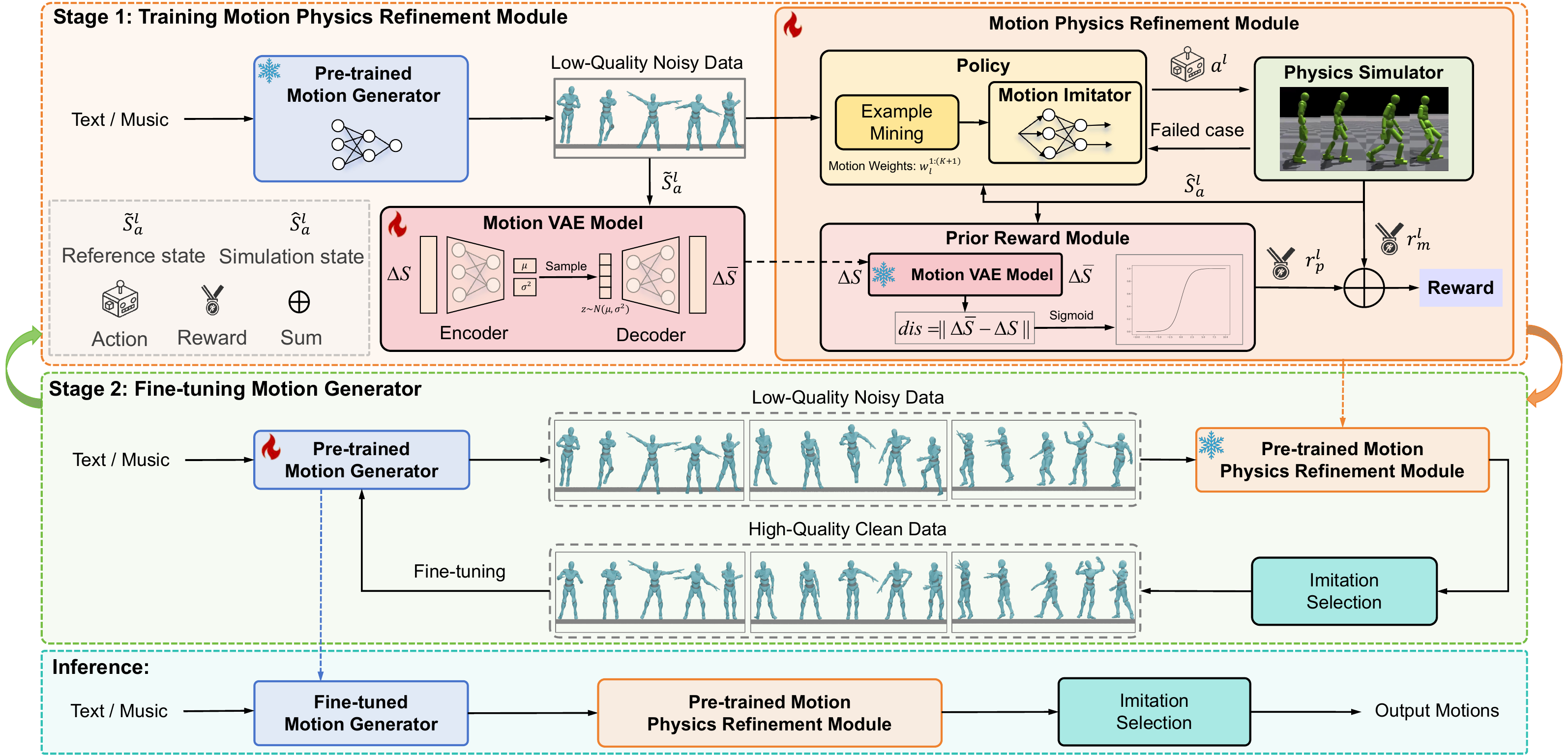}
    \caption{An overview of the Morph framework. Morph comprises a Motion Generator and a Motion Physics Refinement module. Morph employs a two-stage training process: Motion Physics Refinement module training and Motion Generator fine-tuning. And a Imitation Selection Operation is employed to ensure the motion quality after physics refinement. The solid curved arrows on the left and right (in orange and green) represent the iterative, collaborative optimization between Stage 1 and Stage 2.
    }
    \vspace{-0.3cm}
\label{fig:morph_whole_framework}
\end{figure*}

\vspace{0.1cm}
\noindent
\textbf{Physically Plausible Motion Generation.}
Physical plausibility refers to the degree to which generated motions adhere to physical rules, such as foot sliding, foot-ground contact, and body leaning. Generating physically-plausible motions is crucial for many real-world applications, such as animation and virtual reality. 
Recently, a few works have attempted to address this challenge \cite{Yuan_2023_physdiff, han2024reindiffuse, tseng2023edge}. For example,  \cite{Yuan_2023_physdiff,han2024reindiffuse} integrate physical constraints into the diffusion process, iteratively applying physics and diffusion  to maintain alignment with motion data distribution while enhancing physical realism. 
EDGE \cite{tseng2023edge} introduces auxiliary losses to align specific  aspects of physical realism, specifically for generating more physically plausible dances.
However, these methods rely on real motion data and have not fully demonstrated generalizability across different motion generation models and tasks, limiting their flexibility and scalability.

%% file: sections/3_method.tex
\vspace{-0.2cm}
\section{Method}
\vspace{-0.1cm}
To enhance physical plausibility of motion generation without relying on real motion data, we propose a motion-free physics optimization framework, namely morph. As illustrated in Fig.~\ref{fig:morph_whole_framework}, morph consists of two key modules: a Motion Generator (MG) that could be any existing pre-trained motion generation model, and a Motion Physics Refinement module (MPR) independent of Motion Generator. Morph employs a two-stage training process. In the first stage, the Motion Generator produces large-scale noisy motion data (Sec.~\ref{sec3.1}), which is subsequently used to train the Physics Refinement module to map input motions into a physically plausible space (Sec.~\ref{sec3.2}). In the second stage, these physics-refined motions, in turn, are used to fine-tune Motion Generator, further enhancing its capabilities (Sec \ref{sec3.3}). Through these two-stage training process, morph drastically reduces physical errors while improving the overall quality of generated motions.
\subsection{Noisy Motion Data Generation} \label{sec3.1}
Different from ~\cite{Yuan_2023_physdiff, han2024reindiffuse} that rely on real motion data, our goal is to develop a \textit{motion-free}  physical optimization framework. To achieve this, we utilize an existing pretrained motion generation model to produce large-scale, noisy motion data for training the physics refinement module. As shown in Fig.~\ref{fig:morph_whole_framework},  given control signals $c$ (\textit{e.g.}, text or music), the pre-trained motion generator is employed to generate a motion sequence $\boldsymbol{\tilde{x}}^{1:L}$,  as follows:
\begin{equation}
\boldsymbol{\tilde{x}}^{1:L} = f_{\xi}\left(c\right), \text{where}, \boldsymbol{\tilde{x}}^{1:L}=\left\{\boldsymbol{\tilde{x}}^{l}=\left[\boldsymbol{\theta}^l, \boldsymbol{p}^l\right]\right\}_{l=1}^L,
\end{equation}where $f_{\xi}$ represents the motion generator with parameters $\xi$,
$\boldsymbol{\tilde{x}}^l$ denotes the $l^{th}$ pose of the synthetic motion sequence, represented by the joint rotations $\boldsymbol{\theta}$ and positions $\boldsymbol{p}$.  Notably, our framework, morph, is agnostic to the specific instantiation of $f_{\xi}$, allowing it to be generally compatible with various pretrained motion generators.

\subsection{Physics-Based Motion Refinement} \label{sec3.2}
Since most existing motion generation models lack explicit physical constraints, they often produce motions with  noticeable  artifacts, as shown in Fig.~\ref{fig:physic_plaus}. To address this, the Motion Physics Refinement (MPR) module is tasked with projecting  the generated motion $\boldsymbol{\tilde{x}}^{1:L}$, which disregards the laws of physics, into a physically-plausible motion $\boldsymbol{\hat{x}}^{1:L}$. As shown in Fig.~\ref{fig:morph_whole_framework}, the MPR module consists of three components: a motion imitator, a physics simulator and a motion prior reward model. Specifically, the motion imitator controls a simulated character to mimic the input motion  $\boldsymbol{\tilde{x}}^{1:L}$ within the physics simulator. The resulting motion  $\boldsymbol{\hat{x}}^{1:L}$ from the physics simulator is considered physically plausible, as it adheres to the laws of physics. Additionally, the prior reward model penalize the non-human motions further ensuring the naturalness and realism of the physics-refined motions.
\vspace{-0.4cm}
\noindent
\paragraph{Motion Imitator Learning.} \
Motion Imitator Learning can be formulated as a Markov decision process, represented by the tuple $\left(\mathcal{S}, \mathcal{A}, \mathcal{T}, \mathcal{R}\right)$ of states, actions, transition dynamics and reward function.
Formally, the motion imitator is described by a policy $\pi\left(\boldsymbol{a}^l|\boldsymbol{s}^l\right)$, which specifies the probability distribution of selecting an action $\boldsymbol{a}^l \in \mathcal{A}$ given the current state $\boldsymbol{s}^l\in\mathcal{S}$. The physic simulator, in turn, defines the transition dynamics $ \mathcal{T}\left(\boldsymbol{s}^{l+1}|\boldsymbol{s}^l, \boldsymbol{a}^l\right)$, which determines the next state $\boldsymbol{s}^{l+1}$ based on current state $\boldsymbol{s}^l$ and action $\boldsymbol{a}^l$. 
Specifically, starting from an initial state $\boldsymbol{s}^1$, a character agent acts in the physic simulator according to policy  $\pi\left(\boldsymbol{a}^l|\boldsymbol{s}^l\right)$, iteratively sampling action  $\boldsymbol{a}^l$. Then, the physic simulator, governed by the  transition dynamics $ \mathcal{T}\left(\boldsymbol{s}^{l+1}|\boldsymbol{s}^l, \boldsymbol{a}^l\right)$, generates the next state $\boldsymbol{s}^{l+1}$, from which the simulated pose $\boldsymbol{\hat{x}}^{l+1}$ is derived. By running the policy for $L$ steps, we can obtain the simulated motion sequence $\boldsymbol{\hat{x}}^{1:L}$. In implementation, we adopt a standard RL algorithm Proximal Policy Optimization (PPO) to train the motion imitator $\pi$, where a reward is assigned based on how well the simulated motion $\boldsymbol{\hat{x}}^{1:L}$ aligns with input motion $\boldsymbol{\tilde{x}}^{1:L}$. In the following, we elaborate on the design of states, rewards, policy and actions. 

\textit{\textbf{States.}} The simulation state $\boldsymbol{s}^l$ consists of the input motion's next pose,  along with the differences between the input \textit{next}  pose and the \textit{current} simulated pose  across multiple aspects, including joint rotation, position, velocity and angular velocity.
 This difference information informs the policy of the pose residuals that require compensation, enabling the system to better align the simulated motion with the input motion.
Formally, the state  is defined as $\boldsymbol{s}^l = \left[\boldsymbol{\tilde{\theta}}^{l+1}, \boldsymbol{\tilde{p}}^{l+1}, \boldsymbol{\tilde{\theta}}^{l+1} - \boldsymbol{\hat{\theta}}^l, \boldsymbol{\tilde{p}}^{l+1} - \boldsymbol{\hat{p}}^l, \boldsymbol{\tilde{v}}^{l+1} - \boldsymbol{\hat{v}}^l, \boldsymbol{\tilde{\omega}}^{l} - \boldsymbol{\hat{\omega}}^l\right]$. Here, $\boldsymbol{\theta}$, $\boldsymbol{p}$, $\boldsymbol{v}$ and $\boldsymbol{\omega}$ represent the joint rotation, position, joint velocity, and angular velocity, respectively, $\widetilde{\left[\boldsymbol{\cdot}\right]}/\widehat{\left[\boldsymbol{\cdot}\right]}$ denote the quantities of the input$/$simulated motions, and $\left[\boldsymbol{\cdot}\right]^l$ denotes the  quantities at the $l^{th}$ timestep.

\textit{\textbf{Actions.}} 
We use the target joint angles of proportional derivative (PD) controllers as the action representation, where the action $\boldsymbol{a}^l$ specifies the PD target to enable robust motion imitation. 

\textit{\textbf{Policy.}} Following \cite{Yuan_2023_physdiff, han2024reindiffuse, Luo2023PerpetualHC}, we employ a parameterized Gaussian policy $\pi\left(\boldsymbol{a}^l|\boldsymbol{s}^l\right)=\mathcal{N}\left(\mu_{\phi}\left(\boldsymbol{s}^l\right), \Sigma\right)$, where the mean action $\mu_{\phi}\left(\boldsymbol{s}^l\right)$  is the output by our motion imitator, a simple multi-layer perceptron network with parameter $\phi$, and $\Sigma$ is a fixed diagonal covariance matrix. 

\textit{\textbf{Rewards.}} The reward function is designed to encourage the simulated motion  to match the input motion. At each timestep $l$, the reward $r^l$ consists of a mimic reward \(r^l_{\mathrm{m}}\) and a energy penalty \(r^l_{\mathrm{e}}\) and a prior reward $r^l_{\mathrm{p}}$, formulating as $r^l=r^l_{\mathrm{m}}+r^l_{\mathrm{e}}+r^l_{\mathrm{p}}$. The mimic reward \(r^l_{\mathrm{m}}\) includes joint rotation $\boldsymbol{\theta}$, position $\boldsymbol{p}$, joint velocity $\boldsymbol{v}$ and angular velocity $\boldsymbol{\omega}$, as follows:

\vspace{-3mm}
\begin{small}
\begin{equation} \label{eq2}
\begin{aligned}
    &r^l_{\mathrm{m}}= w_{\theta} \exp\left[{-\alpha_\theta\left|\boldsymbol{\tilde{\theta}}^l - \boldsymbol{\hat{\theta}}^l\right|}\right] +  w_{p} \exp\left[{-\alpha_p\left|\boldsymbol{\tilde{p}}^l - \boldsymbol{\hat{p}}^l\right|}\right]  \\
    &+ w_{v} \exp\left[{-\alpha_v\left|\boldsymbol{\tilde{v}}^l - \boldsymbol{\hat{v}}^l\right|}\right] +  w_{\omega} \exp\left[{-\alpha_\omega\left|\boldsymbol{\tilde{\omega}}^l - \boldsymbol{\hat{\omega}}^l\right|}\right], 
\end{aligned}
\end{equation} 
\end{small}where 
$w_{\theta}$, $w_{p}$, $w_{v}$, $w_{\omega}$,   are weighting factors, $\alpha_\theta$, $\alpha_p$, $\alpha_v$,  $\alpha_\omega$ are scaling factors, $\widetilde{\left[\boldsymbol{\cdot}\right]}/\widehat{\left[\boldsymbol{\cdot}\right]}$ denote the quantities of the input$/$simulated pose, and $\left|\cdot\right|$ is the L1 norm.  

The energy  penalty is computed as:
\vspace{-0.1cm}
\begin{equation}
r^l_{\mathrm{e}} = -0.0005 \cdot  \left\|\boldsymbol{\hat{\nu}}^l \boldsymbol{\hat{\omega}}^l\right\|_2^2,
\end{equation}where 
$\boldsymbol{\hat{\nu}}^l$ and $\boldsymbol{\hat{\omega}}^l$ denotes the joint torque and angular velocity of the simulated pose in the $l^{th}$ timestep, and $\left\|\cdot\right\|_2$ is the L2 norm.

\vspace{-0.3cm}
\paragraph{Prior Reward Module.} 
Relying solely on physical simulators and imitation learning often results in non-human-like actions, characterized by a mechanical feel and unrealistic rotations. Previous works \cite{Peng_AMP_2021, Luo2023PerpetualHC} have attempted to address these issues by employing adversarial-based methods trained on high-quality motion data to incorporate human motion knowledge, improving the naturalness of the motion. However, this adversarial-based method often suffers from training instability and produces rewards that are not smooth enough, which also prolongs the overall imitation learning training process. To address this issue, we propose a Motion VAE Reward model as a prior to constrain the motion generated by the physics simulator, making it more natural. As shown in Fig.~\ref{fig:morph_whole_framework}, to accelerate the training speed of motion imitation learning, we first train a lightweight Motion VAE model using motion data generated by the motion generator. This allows the model to effectively learn the distribution of the generated data. The objective function for optimizing the Motion VAE model is as follows:

\vspace{-0.3cm}
\begin{small}
\begin{equation} \label{vae_loss}
\mathcal{L}(\theta, \phi; \mathbf{s}) = 
\mathbb{E}_{q_\phi(\mathbf{z}|\mathbf{\Delta s})} \left[ \log p_\theta(\mathbf{\Delta s}|\mathbf{z}) \right] - D_{\text{KL}} \left( q_\phi(\mathbf{z}|\mathbf{\Delta s}) \parallel p(\mathbf{z}) \right)
\end{equation}
\end{small}where $\theta$ and $\phi$ denote the parameters of VAE, $\Delta s$ refer to the difference in state between two consecutive time steps for reference motion. $D_{KL}(\cdot,\cdot)$ refers to the KL divergence. 

As shown in Fig.~\ref{fig:morph_whole_framework}, after the Motion VAE model is trained, the pre-trained Motion VAE model is applied to compute rewards as follows:
\vspace{-0.1cm}
\begin{small}
\begin{equation} \label{vae_reward}
{r}_{p}^{l} = 1-\frac{1}{1+e^{-||\Delta \overline{s}-\Delta s||}},
\end{equation}
\end{small}
where $\Delta s$ and $\Delta \overline{s}$ refer to the difference in state between two consecutive time steps for simulated motion and the reconstructed simulated motion, respectively. $\left\|\cdot\right\|$ is the L1 norm.



\paragraph{Initialization and Early Termination.} Similar to \cite{Peng_deepmimic}, a starting point is randomly selected from a motion clip for imitation. To improve training efficiency and accelerate convergence, we terminate the episode when the  MPJPE (Mean Per Joint Position Error) between the state of input pose and that of simulated pose exceeds $0.5$ meter.

\vspace{-0.3cm}
\noindent
\paragraph{Hard Negative Mining.} 
As training progresses, the motion imitator gradually learns to imitate simple motion sequences. However, more challenging  examples in the large-scale motion dataset may be overlooked, limiting the model's ability to handle difficult samples. To address this, we implement a Hard Negative Mining process that identifies motions where the  physical simulator fails to imitate as hard samples. Specifically, a \textit{dynamic} weight is assigned to each motion sequence for sampling, doubling whenever imitation fails. This process progressively  increases the focus on challenging samples, guiding the imitator to effectively learn from difficult examples.

\subsection{Motion Generator Fine-tuning} \label{sec3.3} 

Existing motion generators are often trained with limited real motion data, which may limit their capabilities. However, collecting large-scale real 3D motion data is costly. So, we leverage a large volume of synthetic, physically-plausible motion data provided by MPR module, to further enhance the capabilities of the motion generator.


As shown in Fig.~\ref{fig:morph_whole_framework}, in the second stage, we further finetune the motion generator using the physics-refined motions produced by the MPR module. 
Notably, since the physical simulator cannot replicate non-grounded motions (\textit{e.g.}, sitting on a chair or swimming), such simulated motions may deviate from the true data distribution. 
To this end, we apply an \textit{Imitation Selection Operation} to filter out  simulated data of non-grounded motions. Specifically, we calculate the average per-joint position error (MPJPE) between the samples before and after physical optimization.  A threshold $\tau$ is set to determine whether to accept the physically refined motion $\boldsymbol{\hat{x}}^{1:L}$ (with $\text{MPJPE}<\tau$) or input motions $\boldsymbol{\tilde{x}}^{1:L}$ (with $\text{MPJPE}>\tau$).  The selected data is then paired with the original condition signals (\textit{e.g.}, text or music). Ultimately,  a large-scale, physically plausible motion data is constructed for fine-tuning the motion generator. 

Following \cite{guo2024momask,tevet2023human},
we use the mean squared error to optimize the motion generator. Denote the selected motion sequence paired with the  condition as $\left(\boldsymbol{x}^{1:L}, c\right)$,  the objective function for motion generator $f_{\xi}$ is defined as:
\begin{equation}
\mathcal{L}_{\text {MG}}\left(\xi\right)=\mathbb{E}\left[\left\|\boldsymbol{x}^{1:L}-f_{\xi}\left(c\right)\right\|_2^2\right]
\end{equation}
where $\xi$ is the parameters of the motion generator.

\vspace{-0.4cm}
\noindent
\paragraph{Multi-Round Iterative Optimization.} In Fig.~\ref{fig:morph_whole_framework}, we define the complete two-stage process as a single training round. To maximize the synergy between the MG and MPR modules, we repeat this process multiple times, using the model weights from each round as initialization for the next, thus creating a multi-round iterative training paradigm. We present the results of this iterative training in the Appendix, demonstrating its effectiveness and potential.

\vspace{-0.3cm}
\noindent
\paragraph{Inference.} As shown in Fig~\ref{fig:morph_whole_framework}, after the two-stage optimization, the finetuned motion generator (Stage 2) and the trained MPR module (Stage 1) are combined to perform inference. To mitigate simulation error in non-grounded motions, the \textit{imitation selection operation} described above is used as a post-process step. 

%% file: sections/4_experiment.tex
\section{Experiment}
Extensive experiments evaluate the performance of our Morph across multiple motion generation tasks and datasets. Specifically, we assess  Morph on two motion generation tasks, text-to-motion and music-to-dance. As Morph is agnostic to specific instantiation of generation models, we combine Morph with three types of generation models for text-to-motion, \textit{i.e.}, diffusion-based models, including MotionDiffuse \cite{zhang2023t2m} and MDM \cite{tevet2022mdm}, autoregressive-based models, including T2M-GPT \cite{zhang2023t2m} and generative make modeling, including MoMask \cite{guo2024momask}.
For music-to-dance, we combine Morph with the diffusion-based model EDGE \cite{tseng2023edge} and autoregressive model Bailando \cite{siyao2022bailando}. 
The Appendix provides additional visualizations and experimental results showcasing Morph's performance on varying amounts of noisy motion data, different and threshold $\tau$ in imitation selection opeartion, and multi-round optimization of the MPR module and motion generator.

\begin{table*}[!t]
    \centering
    \caption{Ablation study on Morph-MoMask (combined with MoMask \cite{guo2024momask} generator)  for text-to-motion task on HumanML3D dataset. IS: imitation selection operation; Prior: using prior reward training MPR module; Adv: using adversarial reward training MPR module; Energy: using energy reward training MPR module; Real Data: using real motion data training MPR module; FT: fine-tuning Motion Generator with physics-refined motions. The arrows ($\uparrow/\downarrow$) indicate that higher/smaller values are better.
    } \label{ablation_study}
     \vspace{-0.3cm}
     \resizebox{\textwidth}{!}{
    \begin{tabular}{@{}cccccccccccccc@{}}
    \hline
    \specialrule{0em}{1pt}{1pt}
    \multirow{2}{*}{Model  Number} & \multicolumn{6}{c}{Methods} & \multicolumn{2}{c}{Common Generation Metrics} & \multicolumn{5}{c}{Physical Plausibility Metrics} \\ 
    \cmidrule(lr){2-7} \cmidrule(lr){8-9} \cmidrule(lr){10-14}
    & IS & Prior & Adv & Energy & Real Data & FT  & {RTOP-3 $\uparrow$} & {FID $\downarrow$} & {PFC $\downarrow$} & {Penetrate $\downarrow$} & {Float $\downarrow$} & {Skate $\downarrow$} & {IFR $\downarrow$} \\
    \specialrule{0em}{1pt}{1pt} \hline \specialrule{0em}{1pt}{1pt}
        $\mathrm{A}$ & \multicolumn{6}{c}{Baseline (only motion generator MoMask \cite{guo2024momask})} & 0.807 & 0.045 & 1.058 & 23.152 & 10.660 & 5.262 & - \\ \hline \specialrule{0em}{1pt}{1pt}
        $\mathrm{B}$ & & \checkmark & & \checkmark & & & 0.800 & 0.172 & 0.821 & 0.000 & 2.233 & 0.016 & 0.0152 \\
        $\mathrm{C}$ & &  & \checkmark & \checkmark & & & 0.792 & 0.194 & 0.852 & 0.000 & 2.272 & 0.020 & 0.0155 \\
        $\mathrm{D}$ & \checkmark &  & & \checkmark & &  & 0.782 & 0.276 & 0.735 & 0.000 & 2.554 & 0.032 & 0.0362 \\ 
        $\mathrm{E}$ & \checkmark & \checkmark & &  & &   & 0.795 & 0.163 & 0.711 & 0.000 & 2.362 & 0.024 & 0.0270 \\
        $\mathrm{F}$ & \checkmark &  \checkmark & & \checkmark &  &   & 0.802 &0.074 & 0.669 & 0.000 & 2.268 & 0.011 & 0.0153 \\ 
        \hline \specialrule{0em}{1pt}{1pt}
        $\mathrm{G}$ & \checkmark & \checkmark &  & \checkmark & \checkmark &   &  0.792 & 0.179 & 0.734 & 0.000 & 2.420 & 0.015 & 0.0327 \\
        $\mathrm{H}$ & \checkmark & \checkmark & & \checkmark &  &  \checkmark & \textbf{0.816} & \textbf{0.041} & \textbf{0.647} & \textbf{0.000} & \textbf{2.141} & \textbf{0.010} & \textbf{0.0149} \\ 
        \hline
    \end{tabular}
    } \vspace{-0.3cm}
\end{table*}

\subsection{Datasets and Evaluation Metrics}

\noindent
\textbf{HumanML3D.} HumanML3D \cite{Guo_2022_CVPR} is a large-scale 3D human motion-language dataset, consisting of 
14,616 motion clips and 44,970 motion descriptions annotations. The total motion duration is $28.59$ hours, with each motion clip downsampled to 20 FPS and accompanied by 3-4 textual descriptions.  The dataset is split into training, validation and test sets in an $80\%$, $5\%$ and $15\%$ ratios. 

\noindent
\textbf{AIST++.} AIST++ \cite{li2021learn} is a human dance-music dataset, which includes $992$ pieces of high-quality dance sequence across ten dance genres, each dance paired with a corresponding music. The dance durations range from $7.4$ to $48$ seconds, with a frame rate of 60 FPS. The dataset is split into training and evaluation sets, with allocations of $952$ and $40$, respectively.

\noindent
\textbf{Evaluation Metrics.} We evaluate motion generation from two perspectives: common generation metrics and physical plausibility metrics. First, for text-to-motion generation, we follow \cite{guo2022generating, zhang2024motiondiffuse} and use three standard metrics: \textit{Frechet Inception Distance (FID)}, which measures the distance between generated and ground-truth motion distributions; \textit{R-Precision} (RTOP-1, RTOP-2, RTOP-3), which computes retrieval accuracy of generated motions based on input text; and \textit{Diversity (Div)}, which measures the diversity of generated motions. For music-to-dance generation, following \cite{li2024lodge, yu2023long, siyao2022bailando}, we use five metrics: \textit{FID$_k$} and \textit{Div$_k$} for kinetic features (denoted ‘$k$’); \textit{FID$_g$} and \textit{Div$_g$} for geometric features (denoted ‘$g$’); and \textit{Beat Align Score (BAS)}, which measures the alignment between input music and generated dances.

Next, we present the physical plausibility metrics. Following \cite{Yuan_2023_physdiff, tseng2023edge}, we use four physic-based metrics to assess the physical plausibility of generated motions: \textit{Penetrate} that measures ground penetration; \textit{Float} that measures floating; \textit{Skate} that measures foot sliding; \textit{Physical Foot Contact score (PFC)} that measures the realism of foot-ground contact. 
Additionally, we introduce an imitation selection metric, \textit{imitation failure rate (IFR)}, which calculates the failure rate of the physics refinement module in imitating motion.
\vspace{-0.1cm}
\subsection{Implementation Details}
\noindent
\textbf{Training Setup.} We implement Morph based on PyTorch. The NVIDIA's Isaac Gym is used as the physics simulator. During the MPR module's training phase (Stage 1), we employ the Adam optimizer, with a batch size of 64 and a learning rate of $4\times 10^{-5}$. During the finetuning of motion generator (Stage 2), we follow the original training setup, only adjusting  the learning rate to $1 \times 10^{-5}$.
For hyper-parameters setting, we set  $w_{\theta}$, $w_{q}$, $w_v$ and $w_w$ (Eq.~\ref{eq2}) to $0.5$, $0.3$, $0.1$ and $0.1$, $\alpha_{\theta}$, $\alpha_{q}$, $\alpha_v$ and $\alpha_w$ (Eq.~\ref{eq2}) to  $100$, $10$, $0.1$ and $0.1$ and the imitation selection threshold $\tau$ is set to $0.5$.
All our experiments are conducted on 8 Tesla V100 GPUs.
Further details on implementation and hyperparameter analysis are provided in the Appendix.

\noindent
\textbf{Data Preprocessing. }
In text-to-motion generation, some motion sequences involve environmental interactions, such as sitting down, climbing stairs, and swimming, which cannot be simulated by current physical simulator. Therefore, we annotate each text description in HumanML3D \cite{Guo_2022_CVPR} based on its semantic content. Text Descriptions involving interaction-based motions are labeled as `$0$', while other descriptions are labeled as `$1$'. Text labeled as `$1$' will be used to train the MPR module in Stage 1. 
We will release this annotation on HumanML3D in the final version.

\begin{table*}[t]
    \centering
    \captionsetup{font={small}}
    
    \begin{minipage}{\textwidth}
        \centering
    \caption{Comparison results  for text-to-motion task on HumanML3D dataset. Morph is combined with different types of motion generators. MG: Motion Generator; MPR: Motion Physics Refinement module; FT:  fine-tuning motion generator with the physics-refined motion data.  $\dagger$ denotes Morph without fine-tuning the motion generator (only Stage 1 training)
    } \label{results_for_different_generation_models}
    \vspace{-0.3cm}
    \resizebox{\textwidth}{!}{
    \begin{tabular}{@{}lccccccccccc@{}}
        \hline \specialrule{0em}{1pt}{1pt}
        \multirow{2}{*}{MG} & \multirow{2}{*}{MPR} & \multirow{2}{*}{FT} &  \multicolumn{4}{c}{Common Generation Metrics} &  \multicolumn{5}{c}{Physical Plausibility Metrics} \\
        \cmidrule(lr){4-7} \cmidrule(lr){8-12}  
        & & &  RTOP-1 $\uparrow$ & RTOP-3 $\uparrow$ & FID $\downarrow$ & Diversity $\uparrow$ & PFC $\downarrow$ & Penetrate $\downarrow$ & Float $\downarrow$ & Skate $\downarrow$ & IFR $\downarrow$ \\ \hline 
        \specialrule{0em}{1pt}{1pt}
        PhysDiff w/ MD \cite{Yuan_2023_physdiff} &-& -& - & 0.780 & 0.551 & - & - & 0.898 & 1.368 & 0.423 & -  \\
        PhysDiff w/ MDM \cite{Yuan_2023_physdiff} &-&- & - & 0.631 & 0.433 & - & - & 0.998 & 2.601 & 0.512 & - \\ 
        Reindiffuse \cite{han2024reindiffuse} &-&-  & - & 0.622 & 0.385 & - & - & \underline{0.000} & \textbf{0.711} & 0.058 & - \\ \hline \specialrule{0em}{1pt}{1pt}
        MDM \cite{tevet2022mdm} & - & - & 0.455 & 0.749 & 0.489 & \textbf{9.920} & 0.811 & 17.384 & 17.502 & 3.540 & - \\
        Morph-MDM$\dagger$ & \checkmark &  & 0.449 & 0.748 & 0.496 & 9.736 &  0.715 & 0.000 & 2.271 & 0.017 & 0.0176  \\
        Morph-MDM & \checkmark & \checkmark & 0.471 & 0.757 & 0.482 & \underline{9.865} & 0.697 & 0.000 & 2.258 & 0.016 & 0.0153 \\ \hline \specialrule{0em}{1pt}{1pt}
        MotionDiffuse(MD) \cite{zhang2023t2m} & -& -& 0.491 & 0.782 & 0.630 & 9.410 & 0.533 & 16.670 & 5.698 & 3.419 & -  \\ 
        Morph-MD$\dagger$ & \checkmark &   & 0.488 & 0.782 & 0.649 & 9.523 &  \underline{0.433} & 0.000 & 1.162 & 0.028 & 0.0181  \\
        Morph-MD & \checkmark & \checkmark & 0.494 & 0.788 & 0.568 & 9.493 & \textbf{0.414} & 0.000 & \underline{0.853} & 0.029 & 0.0163
        \\  \hline \specialrule{0em}{1pt}{1pt}
        T2M-GPT \cite{zhang2023t2m} &-&- & 0.491  & 0.775 & 0.116 & 9.761 & 0.998 & 72.250 & 8.918 & 7.801 & -  \\ 
        Morph-T2M-GPT$\dagger$ & \checkmark& & 0.490 & 0.774 & 0.133 & 9.648 & 0.764 & 0.000 & 2.702 & 0.042 & 0.0205  \\
        Morph-T2M-GPT & \checkmark & \checkmark & 0.498 & 0.784 & 0.105 & 9.771 & 0.748 & 0.000 & 2.700 & 0.039 & 0.0195 \\ \hline \specialrule{0em}{1pt}{1pt}
        MoMask \cite{guo2024momask} &-&- & \underline{0.521}  & \underline{0.807} & \underline{0.045} & 9.641 & 1.058 & 23.152 & 10.660 & 5.262 & -  \\ 
        Morph-MoMask$\dagger$ & \checkmark& & 0.516 & 0.802 & 0.074 & 9.578 & 0.669 & 0.000 & 2.268 & \underline{0.011} & 0.0153  \\
        Morph-MoMask & \checkmark & \checkmark & \textbf{0.525} & \textbf{0.816} & \textbf{0.041} & 9.689 & 0.645 & \textbf{0.000} & 2.141 & \textbf{0.010} & 0.0149 \\ \hline
        \specialrule{0em}{1pt}{1pt}
    \end{tabular}} \end{minipage}
\end{table*}

\begin{table*}[htbp]
    \centering
    \small
    \begin{minipage}{\textwidth}
        \centering
        \caption{Comparison results on common generation metrics for text-to-motion on HumanML3D dataset.
        } \label{comparison_for_text2motion}
        \vspace{-0.3cm}
\begin{tabular}{@{}lccccccc@{}}
\hline
\specialrule{0em}{1pt}{1pt}
Methods & RTOP-1 $\uparrow$ & RTOP-2 $\uparrow$ & RTOP-3 $\uparrow$ & FID $\downarrow$ & MM-Dist $\downarrow$ & Diversity $\uparrow$ & MModality $\uparrow$ \\ 
\specialrule{0em}{1pt}{1pt}\hline \specialrule{0em}{1pt}{1pt}
        MDM \cite{tevet2022mdm} & 0.455 & 0.645 & 0.749 & 0.489 &  3.330 & \textbf{9.920} & 2.290 \\
        MotionDiffuse (MD) \cite{zhang2024motiondiffuse} & 0.491 & 0.681 & 0.782 & 0.630 & 3.113 & 9.410 & 1.553 \\
        MLD \cite{chen2023executing} & 0.481 & 0.673 & 0.772 & 0.473 & 3.196 & 9.724 & \underline{2.413} \\
        T2M-GPT \cite{zhang2023t2m} & 0.491 & 0.680 & 0.775 & 0.116 & 3.118 & 9.761 & 1.856 \\
        AttT2M \cite{zhong2023attt2m} & 0.499 & 0.690 & 0.786 & 0.112 & 3.038 & 9.700 & \textbf{2.452} \\
        MMM \cite{pinyoanuntapong2024mmm} & 0.515 & 0.708 & 0.804 & 0.089 & 2.926 & 9.577 & 1.226 \\
        MoMask \cite{guo2024momask} & 0.521 & 0.713 & 0.807 & \underline{0.045} & 2.958 & 9.641 & 1.241 \\
        BAMM \cite{pinyoanuntapong2024bamm} & \textbf{0.525} & \underline{0.720} & \underline{0.814} & 0.055 & \textbf{2.919} & 9.717 & 1.687 \\ \hline
            \specialrule{0em}{1pt}{1pt}
        Morph-MDM & 0.471 & 0.664 & 0.757 & 0.482 & 3.103 & \underline{9.865} & 2.179 \\
        Morph-MD & 0.494 & 0.691 & 0.788 & 0.568 & 3.000 &  9.494 & 1.582\\
        Morph-T2M-GPT & 0.498 & 0.690  & 0.784 & 0.105 & 2.994 & 9.771 & 1.910\\
        Morph-MoMask& \textbf{0.525} & \textbf{0.726} & \textbf{0.816} & \textbf{0.041} & \underline{2.920} & 9.689 & 1.655 \\ \hline
\end{tabular} \vspace{-0.35cm} 
\end{minipage} 
\end{table*}
\begin{table*}[t]
  \centering
  \small
    \caption{Comparison results for music-to-dance on AIST++ dataset. $\dagger$ denotes Morph without fine-tuning motion generator.
    }
     \vspace{-0.2cm}
    \resizebox{\textwidth}{!}{
  \begin{tabular}{@{}lcccccccccccc@{}}
    \hline \specialrule{0em}{1pt}{1pt}
     \multirow{2}{*}{MG} &  \multirow{2}{*}{MPR} &  \multirow{2}{*}{FT} & \multicolumn{5}{c}{Common Generation Metrics} &  \multicolumn{5}{c}{Physical Plausibility Metrics} \\
      \cmidrule(lr){4-8} \cmidrule(lr){9-13}
     & & & FID\(_{k}\) $\downarrow$ & FID\(_{g}\) $\downarrow$ & Div\(_{k}\) $\uparrow$ & Div\(_{g}\) $\uparrow$ & BAS $\uparrow$ & PFC $\downarrow$ & Penetrate $\downarrow$ & Float $\downarrow$ & Skate $\downarrow$  & IFR $\downarrow$ \\
    \hline \specialrule{0em}{1pt}{1pt}
    FACT \cite{li2021ai} & -& - & 35.35 & 22.11 & 5.94 & 6.18  & 0.2209 & - & - & - & - & - \\
    TM2D \cite{gong2023tm2d} & -& - & \textbf{19.01} & 20.09 & \textbf{9.45} & 6.36  & 0.2049 & - & - & - & - & - \\
    EDGE \cite{tseng2023edge}& -& - & 42.16 & 22.12 & 3.96 & 4.61 & 0.2334 & 0.610 & 76.490 & 60.982 & 7.906 & -  \\ 
    Bailando \cite{siyao2022bailando} & -& - & 28.16 & \underline{9.62} & 7.83 & 6.34  & 0.2332 & 0.074 & 31.183 & 12.650 & 4.466 & - \\ \hline \specialrule{0em}{1pt}{1pt}
    Morph-EDGE$\dagger$ & \checkmark &  & 43.01  & 23.69 & 4.30 & 5.18 & 0.2229 & 0.302 & 0.000 & 3.720 & \underline{0.011} & 0.0092\\
    Morph-EDGE & \checkmark & \checkmark & 39.24  & 19.82 & 4.93 & 5.36 & 0.2329 & 0.295 & 0.000 & 3.519 & \textbf{0.009} & 0.0082 \\
    Morph-Bailando$\dagger$  & \checkmark &  & 35.48  & 12.03 & 7.70 & \underline{6.49} & \underline{0.2409} & \underline{0.044} & 0.000 & \underline{2.076} & 0.023 & \underline{0.0064} \\
    Morph-Bailando & \checkmark & \checkmark &  \underline{26.25}  & \textbf{9.38} & \underline{7.88} & \textbf{6.62} & \textbf{0.2412} & \textbf{0.043} & \textbf{0.000} & \textbf{2.057} & 0.021 & \textbf{0.0055} \\
    \hline
  \end{tabular}} \vspace{-0.3cm}
  \label{table_music2dance_results_comparison}
\end{table*}

\vspace{-0.1cm}
\subsection{Ablation Studies}
In this section, we conduct ablation studies to validate the effectiveness of each part of our method. We use MoMask~\cite{guo2024momask} as the motion generator in Morph, denoted as Morph-MoMask, throughout the ablation studies.
The ablation results are shown in Tab.~\ref{ablation_study}. 

\vspace{0.0cm}
\noindent
\textbf{Effectiveness of training MPR module using only generated data.} 
We investigate the effectiveness of using only synthetic motion data from two perspectives: its overall effectiveness and its performance relative to training with real motion data. First, we compare the baseline model $\mathrm{A}$  (only the MoMask generator) and model $\mathrm{F}$ which combines MoMask with the trained MPR module. In Tab.~\ref{ablation_study}, model $\mathrm{F}$ outperforms model $\mathrm{A}$ on all physical metrics ($0.669/1.058$ on PFC, $0.0/23.152$ on Penetrate, $2.268/10.66$ on Float, $0.011/5.262$ on Skate), while maintaining competitive performance on generation metrics. These results confirm the feasibility of training the MPR module solely with synthetic data. 
Next, we compare model $\mathrm{F}$  with model $\mathrm{G}$, which uses real motion data to train the MPR module. In Tab.~\ref{ablation_study}, model $\mathrm{G}$ achieves inferior performance compared to model $\mathrm{F}$.
This can be attributed to the significant domain gap between the training data (real motions) and test data (generated motions). This gap  hinders the MPR module trained on real data from effectively adapting to the generated motion data, leading to performance decline.

\vspace{0.0cm}
\noindent
\textbf{Effectiveness of prior reward and energy reward.} 
When training the MPR module, we add an prior reward and energy reward alongside the common imitation task reward. In Tab.~\ref{ablation_study}, comparing model $\mathrm{F}$ with model $\mathrm{D}$, which omits the prior reward, we observe a performance decline in both generation and physical metrics on model $\mathrm{D}$. The significant drop in FID metric indicates that, without the prior reward, the simulated motions lack the distribution constraints of the input motions, leading to a deviation from the intended  motion distribution. The $\mathrm{B}$, which leverages prior rewards, significantly outperforms model $\mathrm{C}$, which uses adversarial rewards, across various generation and physics-based metrics. This highlights the superior effectiveness of the proposed prior reward over the adversarial reward. We compare model $\mathrm{F}$ with $\mathrm{E}$, which omits the energy reward. As shown in Tab.~\ref{ablation_study}, model $\mathrm{F}$ outperforms model $\mathrm{E}$ in generation metrics. We argue that introducing energy reward helps suppress high-frequency jitter during the humanoid robot control process, resulting in more natural motions.

\vspace{-0.06cm}
\noindent 
\textbf{Effectiveness of imitation selection strategy.} 
Like~\cite{Yuan_2023_physdiff, han2024reindiffuse}, the proposed Morph focuses on addressing physical inconsistencies in non-interactive and ground-contact human motions. In training and testing data, there are some interactive and non-grounded motions, such as sitting on a chair, swimming, or climbing stairs. To ensure the quality of the motions after refining, we design an imitation selection operation to filter out the simulations of such motions. In Tab.~\ref{ablation_study}, we compare model $\mathrm{F}$ with $\mathrm{B}$, which omits the imitation selection operation. As shown, model $\mathrm{F}$ outperforms $\mathrm{B}$ in both generation and physic metrics, validating the effectiveness of the imitation selection strategy.

\vspace{-0.17cm}
\noindent
\textbf{Effectiveness of finetuning motion generator using simulated motions.}
Furthermore, we compare model $\mathrm{F}$ and model $\mathrm{H}$, which finetunes the motion generator using the physically refined motions produced by the MPR module. As shown in Tab.~\ref{ablation_study}, model $\mathrm{H}$ achieves further gains in both generation and physic metrics, with RTOP-3 increasing by $1.4\%$ and FID improving by $0.033$. These results suggest that the proposed MPR module can, in turn, enhance the performance of the motion generator.

\vspace{-0.45cm}
\subsection{Evaluation with Different Motion Generators}
\vspace{-0.15cm}
We evaluate the adaptability of our Morph with different motion generators, including MDM \cite{tevet2022mdm}, Motiondiffuse \cite{zhang2024motiondiffuse}, T2M-GPT \cite{zhang2023t2m}, and MoMask \cite{guo2024momask}. 
As shown in Tab.~\ref{results_for_different_generation_models}, Morph significantly enhances  physical fidelity while maintaining competitive generation metrics across different motion generation models. For example, after integrating Morph with MDM (Morph-MDM), RTOP-1 metric improved by $0.016$. In terms of  physical plausibility metrics, penetration dropped to zero, float decreased from $17.502$ to $2.258$, and skate reduced from $3.540$ to $0.016$. Similar performance improvements are observed with other motion generators.
The consistent gains across different generators highlight the versatility of our Morph framework.

\begin{figure}[t]
	\centering
	\tabcolsep=0.05cm
    \includegraphics[width=0.84\linewidth]{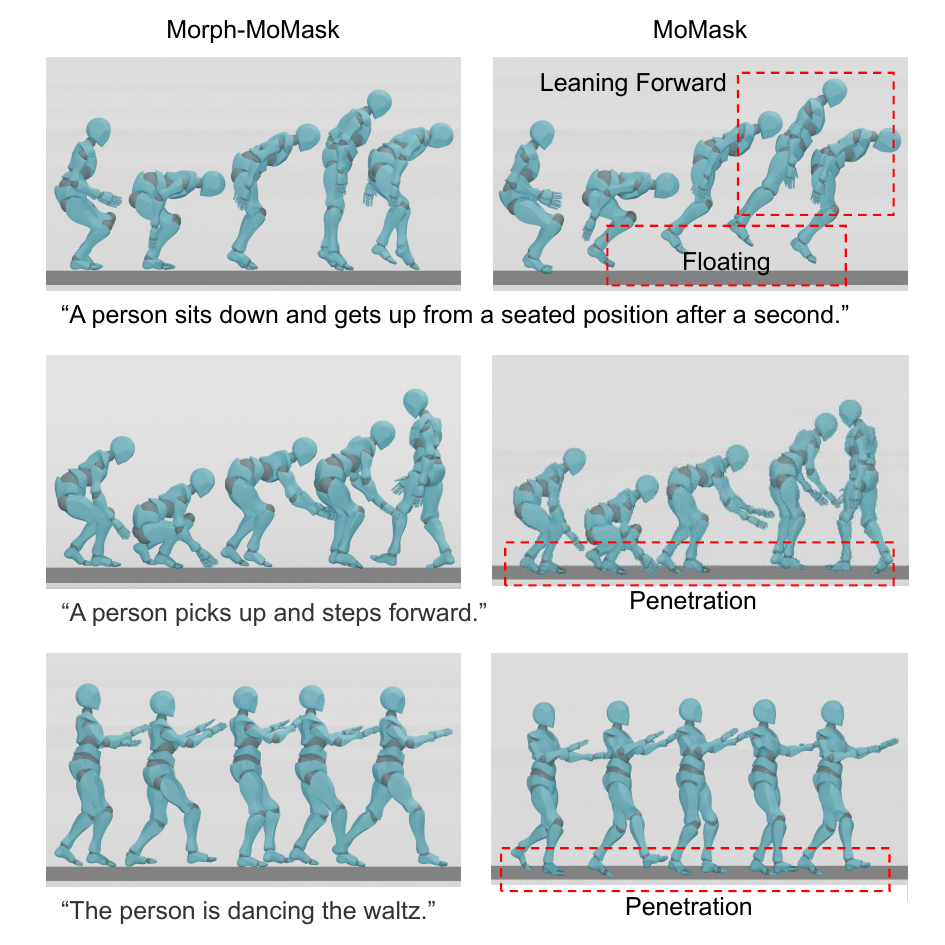}
    \vspace{-0.2cm}
    \caption{Qualitative comparison between our Morph-MoMask and MoMask in text-to-motion task. Morph-MoMask significantly reduces physical artifacts such as leaning forward, floating and penetration. } \label{fig_visualization}
    \vspace{-0.4cm}
\end{figure}

\subsection{Comparisons with State-of-the-arts}
\noindent
\textbf{Main results on Text-to-Motion Generation.} On the text-to-motion dataset, HumanML3D, we compare Morph framework with other  state-of-the-art methods. The comparison results are shown in Tab.~\ref{results_for_different_generation_models} and Tab.~\ref{comparison_for_text2motion}. As shown, Morph achieves significant gains in physical metrics while maintaining competitive performance in generation metrics, demonstrating  its capability to enhance physical plausibility. 
Fig.~\ref{fig_visualization} illustrates  the generated motions from both our Morph-MoMask and MoMask. As shown, motions generated by MoMask often exhibit physically unrealistic artifacts, such as penetration, floating, and leaning forward. In contrast, Morph-MoMask effectively reduces these artifacts, producing motions that are both physically plausible and realistic.  

\noindent
\textbf{Main results on Music-to-Dance Generation.}
On the music-to-dance dataset, AIST++, we compare Morph combined with  EDGE  \cite{tseng2023edge} and Bailando \cite{siyao2022bailando} generators against other state-of-the-art methods. As shown in Tab.~\ref{table_music2dance_results_comparison}, Morph significantly improves physical metrics compared to existing methods. For generation metrics, our Morph-EDGE model achieves the best results in multiple metrics, including FID$_g$, Div$_g$, and BAS, while also demonstrating competitive performance in FID$_k$. These results validate the superiority of our framework in music-to-dance generation. 

%% file: sections/5_conclusion.tex
\vspace{-0.2cm}
\section{Conclusion}
In this paper, we present Morph, a model-agnostic physical optimization framework designed to enhance physical plausibility in motion generation without relying on costly real-world data. We begin by using a pretrained motion generator (MG) to synthesize large-scale noisy motion data. Then, we introduce a Motion Physics Refinement (MPR) module that leverages this synthesized data to train a motion imitator enforcing physical constraints. These physically refined motions are used to fine-tune the motion generator, further enhancing its capabilities. To stabilize the physics optimization process, we introduce a prior reward module. This collaborative training paradigm enables mutual enhancement between MG and MPR module, improving practicality and robustness. Our framework is compatible with various motion generation models for both text-to-motion and music-to-dance tasks. Extensive experiments show that Morph substantially improves physical plausibility while achieving competitive generation quality.

%% file: sections/supplementary_material.tex
\appendix
\section*{\begin{center}
    Appendix
\end{center}} \label{appendix}

\newcounter{appendixsection}
\renewcommand{\theappendixsection}{A\arabic{appendixsection}}
\newcommand{\appendixsection}[1]{%
    \refstepcounter{appendixsection}
    \section*{\theappendixsection\hspace{1em}{#1}}
    \addcontentsline{toc}{section}{\theappendixsection\hspace{1em}{#1}}
}

In this Appendix, we present more details for Morph, including data preprocess, additional experimental results, qualitative comparisons. First, we describe the data preprocessing procedure used for training the Motion Physics Refinement (MPR) module with generated motion data (Sec.~\ref{data_preprocess}). Then, we present experimental results analyzing the impact of $\tau$ in the imitation selection operation (Sec.~\ref{is_threshold}), the effect of varying the quantity of noisy motion data for MPR training (Sec.~\ref{varying_data}), and effect of the number of training rounds for Morph (Sec.~\ref{multi_round}). Finally, we provide additional qualitative comparisons for text-to-motion and music-to-dance tasks (Sec.~\ref{qualitative_comparisons}). 

\section{Details for Data Preprocess} \label{data_preprocess}

\begin{figure}[t]
	\centering
	\tabcolsep=0.05cm
    \begin{minipage}[t]{0.48\textwidth}
    \centering
      \includegraphics[width=1\linewidth]{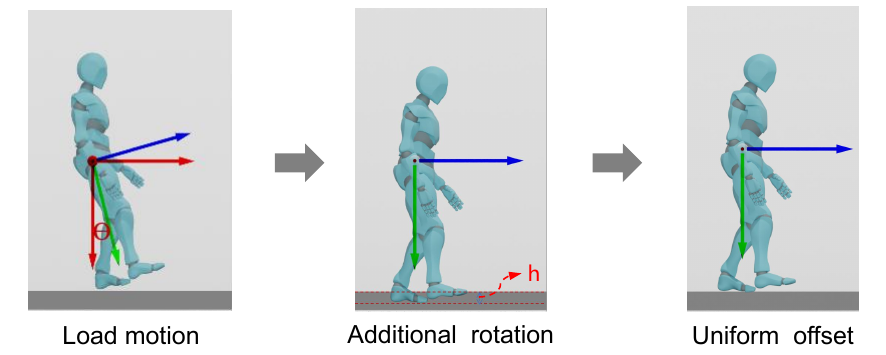}
    \caption{A flowchart illustrating the data preprocessing process. The parameters are calculated from the first frame and then applied to all generated motion sequences before they are fed into the MPR module.}
    \vspace{-0.1cm}
    \label{fig: preprocess}  
    \end{minipage}

\begin{minipage}[t]{0.48\textwidth}
        \centering
        \begin{algorithm}[H]
\caption{Preprocessing Motion Sequences}
\label{alg:preprocess}
\begin{algorithmic}[1]
    \Require Motion sequence $S$ with frames $F_1, F_2, \ldots, F_n$
    \Ensure Preprocessed motion sequence $S'$
    
    \Statex \textbf{Step 1: Calculate the angle $\theta$}
    \begin{enumerate}[label=(\arabic*)]
                \item Compute the projection of the center of mass of $F_1$ onto the ground.
                \item Determine the line connecting the pelvis point and the center of both feet in $F_1$.
                \item Calculate the angle $\theta$ between the projection and the line.
            \end{enumerate}

    \Statex \textbf{Step 2: Correct posture if $\theta > 10^\circ$}
    \begin{enumerate}[label=(\arabic*)]
                \item Apply an additional rotation to the pelvis for the entire sequence $S$.
            \end{enumerate}

    \Statex \textbf{Step 3: Ensure $F_1$ is on the ground}
    \begin{enumerate}[label=(\arabic*)]
                \item Infer the lowest point height $h$ of the mesh in $F_1$.
                \item Add a uniform offset to the entire sequence $S$.
            \end{enumerate}

    \Statex \textbf{Step 4: Output the preprocessed sequence $S'$}.
\end{algorithmic}
\end{algorithm}
\end{minipage}
\end{figure}

As discussed in the main text, the generated motion sequences may exhibit issues such as body leaning, floating and ground penetration. When imported into the simulator, these issues can cause instability in the robot, potentially causing it to fall, bounce off the ground, or drop from mid-air. To address this issue, we apply a preprocessing step to the motion sequences, detailed in  Fig.~\ref{fig: preprocess} and Alg.~\ref{alg:preprocess}. Specifically, we first compute the body's tilt angle, defined as the angle between the projection of the center of mass onto the ground and the line connecting both feet. If this angle exceeds $10^\circ$, we apply the necessary adjustment to the pelvis throughout the sequence. To correct floating and penetration, we determine the lowest mesh height and adjust the entire sequence by this offset. The preprocessed sequence is then used for training and inference.

\begin{table*}[t]
    \centering
    \begin{minipage}{\textwidth}
    \centering
    \caption{Hyper-parameter analysis of $\tau$ in Imitation Selection operation. Comparison with different values of $\tau$ based on Morph-MoMask$\dagger$ (combined with MoMask \cite{guo2024momask} motion generator, without fine-tuning motion generator)  for text-to-motion task on HumanML3D dataset. The arrows ($\uparrow/\downarrow$) indicate that higher/smaller values are better.
    } \label{effect_of_different_threshold_in_is}
     \vspace{-0.3cm}
    \begin{tabular}{@{}cccccccccc@{}}
    \hline
    \specialrule{0em}{1pt}{1pt}
     \multirow{2}{*}{Methods} & \multicolumn{4}{c}{Common Generation Metrics} & \multicolumn{5}{c}{Physical Plausibility Metrics} \\ 
     \cmidrule(lr){2-5} \cmidrule(lr){6-10}
    & {RTOP-1 $\uparrow$} & {RTOP-3 $\uparrow$} & {FID $\downarrow$} & Diversity $\uparrow$ & {PFC $\downarrow$} & {Penetrate $\downarrow$} & {Float $\downarrow$} & {Skate $\downarrow$} & {IFR $\downarrow$} \\
    \specialrule{0em}{1pt}{1pt} \hline \specialrule{0em}{1pt}{1pt}
    $\tau$=0.0  & \textbf{0.521} & \textbf{0.807} & \textbf{0.045} & \textbf{9.641} & 1.058 & 23.152 & 10.660 & 5.262 & - \\ \specialrule{0em}{1pt}{1pt}
    $\tau$=0.1  & 0.520 & 0.806 & 0.048 & 9.636 & 0.877 & 3.564 & 4.779 & 2.128 & 0.1281 \\ \specialrule{0em}{1pt}{1pt}
    $\tau$=0.2  & 0.519 &  0.805 & 0.056 & 9.625 & 0.771 & 0.838 & 4.015 & 1.057 & 0.0540 \\  \specialrule{0em}{1pt}{1pt}
    $\tau$=0.3 & 0.518 &  0.805 & 0.067 & 9.583 & 0.757 & 0.054 & 3.200 & 0.529 & 0.0258 \\ \specialrule{0em}{1pt}{1pt}
    $\tau$=0.4 & 0.517 &  0.803 & 0.071 & 9.584 & 0.722 & 0.002 & 2.991 & 0.211 & 0.0158 \\ \specialrule{0em}{1pt}{1pt}
    \rowcolor{green!40} $\tau$=0.5 & 0.516 &  0.802 & 0.074 & 9.578 & 0.669 & 0.000 & 2.268 & 0.011 & 0.0153 \\  \specialrule{0em}{1pt}{1pt}
    $\tau$=0.6 & 0.512 & 0.801  & 0.079  & 9.576 & 0.664 & 0.000 & 2.263 & 0.010 & 0.0144 \\  \specialrule{0em}{1pt}{1pt}
    $\tau$=0.7 & 0.510 & 0.799 & 0.080 & 9.543 & 0.660 & 0.000 & 2.093 & 0.006 & 0.0128  \\ \specialrule{0em}{1pt}{1pt}
    $\tau$=0.8 & 0.506 & 0.797 & 0.081 & 9.520 & 0.645 & 0.000 & 2.022 & 0.005 & 0.0124  \\ \specialrule{0em}{1pt}{1pt}
    $\tau$=0.9 & 0.504 & 0.795 & 0.085 & 9.408 & 0.634 & 0.000 & 1.985 & 0.004 & 0.0117  \\ \specialrule{0em}{1pt}{1pt}
    $\tau$=1.0 & 0.497 & 0.793 & 0.084 & 9.255 & \textbf{0.623} & \textbf{0.000} & \textbf{1.982} & \textbf{0.003} & \textbf{0.0111}  \\ \specialrule{0em}{1pt}{1pt}
    \hline
    \end{tabular}
    \vspace{0.5cm}
    \end{minipage}
    
\begin{minipage}{\textwidth}
    \centering
    \caption{Comparison of text-to-motion with different amounts of noisy motion data training for Morph-MoMask$\dagger$ (combined with MoMask \cite{guo2024momask} motion generator, without fine-tuning motion generator). $N$ refers to the total number of generated noisy motion data samples, which is three times the amount of the original real training data. $D$ refers to the number of generated motion data used to train the MPR module. We set $\tau$ as 0.5 for testing.
    } \label{effect_of_different_amounts_of_generated_motion}
     \vspace{-0.3cm}
    \begin{tabular}{@{}lccccccccc@{}}
    \hline
    \specialrule{0em}{1pt}{1pt}
     \multirow{2}{*}{Methods} & \multicolumn{4}{c}{Common Generation Metrics} & \multicolumn{5}{c}{Physical Plausibility Metrics} \\ 
     \cmidrule(lr){2-5} \cmidrule(lr){6-10}
    & {RTOP-1 $\uparrow$} & {RTOP-3 $\uparrow$} & {FID $\downarrow$} & Diversity $\uparrow$ & {PFC $\downarrow$} & {Penetrate $\downarrow$} & {Float $\downarrow$} & {Skate $\downarrow$} & {IFR $\downarrow$} \\
    \specialrule{0em}{1pt}{1pt} \hline \specialrule{0em}{1pt}{1pt}
    $D$=25$\%N$  & 0.495 & 0.791 & 0.087 & 9.477 & 0.866 & 0.120 & 2.997 & 0.035 & 0.0262 \\ \specialrule{0em}{1pt}{1pt}
    $D$=50$\%N$  & 0.498 & 0.795 & 0.082 & 9.536 & 0.815 & 0.022 & 2.870 & 0.023 & 0.0205 \\ \specialrule{0em}{1pt}{1pt}
    $D$=75$\%N$  & 0.512 & 0.800 & 0.078 & 9.569 & 0.761 & 0.002 & 2.429 & 0.012 & 0.0178 \\  \specialrule{0em}{1pt}{1pt}
    $D$=100$\%N$ & \textbf{0.516} &  \textbf{0.802} & \textbf{0.074} & \textbf{9.578} & \textbf{0.669} & \textbf{0.000} & \textbf{2.268} & \textbf{0.011} & \textbf{0.0153} \\ \specialrule{0em}{1pt}{1pt}
    \hline
    \end{tabular}
    \vspace{0.5cm}
    \end{minipage}
    
\begin{minipage}{\textwidth}
    \centering
    \caption{Comparison of text-to-motion with multi-round optimization of the MPR module and motion generator based on Morph-MoMask. We set $\tau$ as 0.5 and use the total number of generated noisy motion data to train.
    } \label{effect_of_multi_round_optimization}
     \vspace{-0.3cm}
    \begin{tabular}{@{}lccccccccc@{}}
    \hline
    \specialrule{0em}{1pt}{1pt}
     \multirow{2}{*}{Methods} & \multicolumn{4}{c}{Common Generation Metrics} & \multicolumn{5}{c}{Physical Plausibility Metrics} \\ 
     \cmidrule(lr){2-5} \cmidrule(lr){6-10}
    & {RTOP-1 $\uparrow$} & {RTOP-3 $\uparrow$} & {FID $\downarrow$} & Diversity $\uparrow$ & {PFC $\downarrow$} & {Penetrate $\downarrow$} & {Float $\downarrow$} & {Skate $\downarrow$} & {IFR $\downarrow$} \\
    \specialrule{0em}{1pt}{1pt} \hline \specialrule{0em}{1pt}{1pt}
    One-Round w/o FT  & 0.516 &  0.802 & 0.074 & 9.578 & 0.669 & 0.000 & 2.268 & 0.011 & 0.0153 \\ \specialrule{0em}{1pt}{1pt}
    One-Round  &  0.525 &  0.816 & 0.041 & 9.689 & 0.645 & 0.000 & 2.141 & 0.010 & 0.0149 \\ \hline \specialrule{0em}{1pt}{1pt}
    Two-Round w/o FT  & 0.526 & 0.817 & 0.041 & 9.692 & 0.632 & 0.000 & 2.129 & 0.010 & 0.0134 \\  \specialrule{0em}{1pt}{1pt}
    Two-Round & 0.527 & 0.818 & 0.038 & 9.697 & 0.625 & 0.000 & 2.108 & 0.007 & 0.0129  \\ \specialrule{0em}{1pt}{1pt}
    \hline \specialrule{0em}{1pt}{1pt}
    Three-Round w/o FT  & 0.526 & 0.821 & 0.040 & 9.701 & 0.621 & 0.000 & 2.121 & 0.009 & 0.0131 \\  \specialrule{0em}{1pt}{1pt}
    Three-Round & \textbf{0.528} & \textbf{0.823} & \textbf{0.034} & \textbf{9.715} & \textbf{0.618} & \textbf{0.000} & \textbf{2.100} & \textbf{0.006} & \textbf{0.0122}  \\ 
    \hline
    \end{tabular}
    \end{minipage} \vspace{-0.3cm}
\end{table*}

\section{Effect of $\tau$ in Imitation Selection on Morph} \label{is_threshold}
In Tab.~\ref{effect_of_different_threshold_in_is}, we analyze the effect of the threshold $\tau$ in the imitation selection operation on Morph. Different values of $\tau$ are tested to assess the performance of Morph-MoMask$\dagger$ (combined with MoMask \cite{guo2024momask} motion generator, without fine-tuning motion generator). 
When $\tau$ is set to 0, the motion refined by the MPR module is not utilized, and Morph directly outputs the results from the motion generator. As $\tau$ increases, the physical plausibility metrics improve significantly. However, the generation metrics show a slight decrease due to the inclusion of some incorrectly refined or non-grounded motions at higher thresholds. 
Larger values of $\tau$ incorporate more refined motions, improving the physical plausibility metrics. 
However, this also increases the acceptance of incorrectly refined motions, leading to a shift in the motion distribution and a corresponding decline in the generation metrics. According to Tab.~\ref{effect_of_different_threshold_in_is}, we observe that $\tau = 0.5$ strikes a balance between generation  and physical plausiibility metrics. Therefore,  we set $\tau$ to 0.5 in this paper.

\section{Effect of Varying Amounts of Noisy Motion Data on Morph} \label{varying_data}
In Tab.~\ref{effect_of_different_amounts_of_generated_motion}, we investigate the impact of varying amounts of generated motion data on the training of Morph.  Different numbers of generated motion data are used to train the MPR module in Morph-MoMask$\dagger$. As shown in Tab.~\ref{effect_of_different_amounts_of_generated_motion}, increasing the amount of training data for the Motion Physics Refinement (MPR) module leads to improvements in both the generation and physical plausibility metrics on the test set. 
These results indicate that a larger volume of generated motion data enhances the MPR module's ability to better mimic the input motion and produces higher-quality outputs.
Conversely, when the MPR module is trained with a smaller dataset, its motion imitation capability diminishes, leading to greater discrepancies between the generated and input motions. This results in a decline in both the generation and physical plausibility metrics. 
These results further highlight the effective data augmentation capability of our proposed Morph. 
\begin{figure*}[!t]
	\centering
	\tabcolsep=0.05cm
    \includegraphics[width=0.99\linewidth]{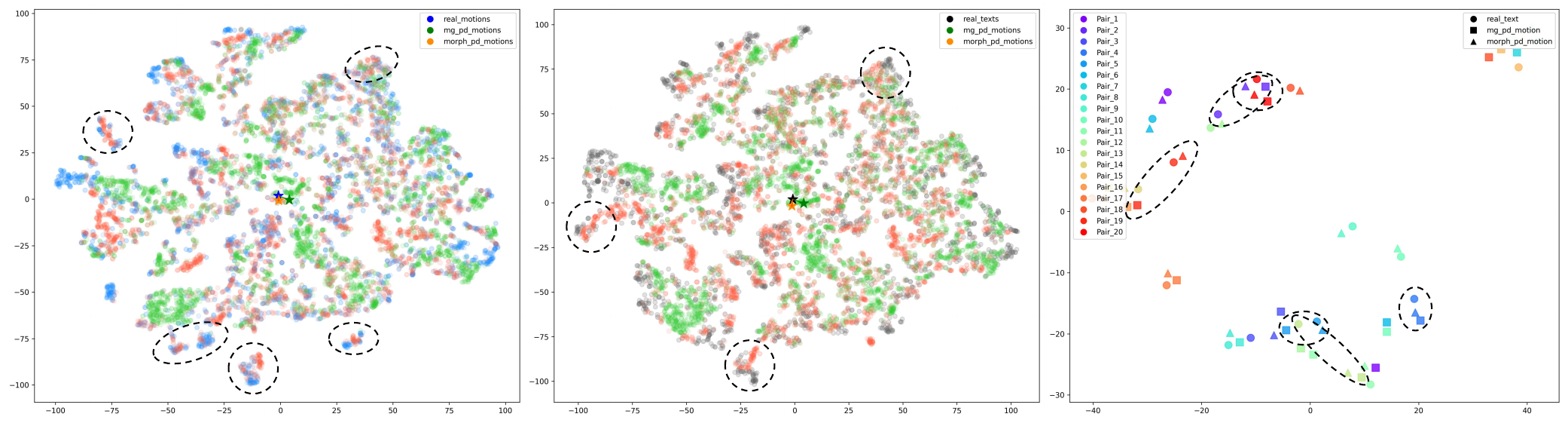}
    \caption{T-sne of motion and text distribution between MG and Morph.} 
\label{fig:tsne}
\end{figure*}

\begin{table}[H]
\centering
\caption{The win rate of Morph over baselines.}
\label{tab:win_rate}
\vspace{-0.2cm}
\resizebox{0.48\textwidth}{!}{
\begin{tabular}{lcccc}
\hline
\specialrule{0em}{1pt}{1pt}
\textbf{Module} & \textbf{Semantic Consistency} & \textbf{Realism} & \textbf{Physical Plausibility} & \textbf{Fluency} \\ \hline \specialrule{0em}{1pt}{1pt}
MDM-Morph vs. MDM  & 84.8\% & 80.1\% & 96.6\% & 85.0\% \\
T2M-GPT-Morph vs. T2M-GPT  & 87.1\% & 79.6\% & 94.4\% & 81.2\%\\
MoMask-Morph vs. MoMask & 90.4\% & 88.3\% & 97.5\% & 80.9\%\\
\specialrule{0em}{1pt}{1pt}
\hline
\end{tabular}}
\end{table}

\section{Effect of Multi-Round Optimization of the MPR module and MG on Morph} \label{multi_round}
In Tab.~\ref{effect_of_multi_round_optimization}, we analyze the effect of multi-round optimization of the  Physics Refinement (MPR) module and Motion Generator (MG) on Morph using Morph-MoMask. To further validate the effectiveness of this round-based training approach in enhancing both the MG and the MPR module, we conducted an additional round of training beyond this single-round training described in the main text. This extra round explores the potential for mutual enhancement between the two modules. In Tab. \ref{effect_of_multi_round_optimization}, the following terms are defined: 
\begin{itemize}
    \item \textit{One-Round w/o FT}: The first round of training where only the MPR module is trained.
    \item  \textit{One-Round}: The first round of training that includes both training the MPR module and fine-tuning the MG.
    \item \textit{Two-Round w/o FT}: Training the MPR module again using the motion data generated by the fine-tuned MG from the first round.
    \item \textit{Two-Round}: Fine-tuning the Motion Generator using the results from \textit{Two-Round w/o FT}.
    \item \textit{Three-Round w/o FT}: Training the MPR module again using the motion data generated by the fine-tuned MG from the second round.
    \item \textit{Three-Round}: Fine-tuning the Motion Generator using the results from \textit{Three-Round w/o FT}.
\end{itemize}

As shown in Tab.~\ref{effect_of_multi_round_optimization}, in the first round of training, MG improves the performance of MPR module, enhancing the physical quality of its generated motion. The refined motion data from the trained MPR module is then used to fine-tune the MG, boosting its performance further. 
In the second round, the fine-tuned MG from the first round is used to generate training data for the MPR module (initialized with first-round weights). 
We observed improvements in \textit{Two-Round w/o FT} compared to \textit{One-Round}, with PFC increasing by 0.013, Float by 0.012, and IFR decreasing, indicating enhanced motion imitation by the MPR module. 
After fine-tuning the MG once again, \textit{Two-Round} shows improvements in the RTOP-1 and RTOP-3 metrics. The model's generation and physical performance reached their best in Three-Round.
These results clearly demonstrate that the MG and MPR modules can mutually enhance each other. Moreover, alternating training between the MG and MPR modules across multiple rounds can further improve the performance of Morph.
\begin{table*}[h]
\centering

\caption{Cross-Task generalization results on Music2Dance and Text2Motion}
\label{tab:crosstask}
\scriptsize
\resizebox{1.0\textwidth}{!}{
\begin{tabular}{@{}lcccccc@{}}
\hline
\specialrule{0em}{1pt}{1pt}
\textbf{Task} & \textbf{$FID / FID_k$} $\downarrow$ & \textbf{$FID_g$} $\downarrow$ & \textbf{$RTOP3 / Div_k$} $\uparrow$  & \textbf{$PFC$} $\downarrow$ & \textbf{$Penetrate$} $\downarrow$ & \textbf{$Float$} $\downarrow$\\
\specialrule{0em}{1pt}{1pt}
\hline
\hline
\specialrule{0em}{1pt}{1pt}
\multicolumn{7}{c}{ Cross-task generalization evaluation} \\
\specialrule{0em}{1pt}{1pt}
\hline
\specialrule{0em}{1pt}{1pt}
\textit{Text-to-Motion (MPR trained on T2M)}  & 0.074 & --  & 0.802 & 0.669  & 0.000 & 2.268 \\
\textit{Text-to-Motion (MPR trained on M2D)}  & 0.116 & --  & 0.795 & 0.881  & 0.000 & 2.436 \\
\hline
\specialrule{0em}{1pt}{1pt}
\textit{Music-to-Dance (MPR trained on M2D)} & 35.48 & 7.70 & 12.03 & 0.044 & 0.000 & 2.076 \\
\textit{Music-to-Dance (MPR trained on T2M)} & 37.66  & 7.38  & 14.02& 0.057 & 0.000 & 2.193 \\
\hline
\hline
\specialrule{0em}{1pt}{1pt}
\multicolumn{7}{c}{Action-to-motion tests} \\
\specialrule{0em}{1pt}{1pt}
\hline
\specialrule{0em}{1pt}{1pt}
\textit{Action\_Label-to-Motion (MDM-action)}  & 0.497& -- & 0.396 &0.544& 15.770 & 7.467 \\
\textit{Action\_Label-to-Motion (MDM-action with Morph)}  & 0.424 & --   & 0.416& 0.509 & 0.000 & 2.115 \\
\hline
\hline
\specialrule{0em}{1pt}{1pt}
\multicolumn{7}{c}{GAN-based Transformer tests} \\
\specialrule{0em}{1pt}{1pt}
\hline
\specialrule{0em}{1pt}{1pt}
\textit{Text-to-Motion (GAN-based Transformer)}  &  0.628 & - & 0.736 & 0.933 & 47.612 & 21.008 \\
\textit{Text-to-Motion (GAN-based Transformer with Morph)}  & 0.606 & - & 0.755 & 0.742 & 0.000 & 2.637 \\
\hline
\hline
\specialrule{0em}{1pt}{1pt}
\multicolumn{7}{c}{Long-duration dance samples tests} \\
\specialrule{0em}{1pt}{1pt}
\hline
\specialrule{0em}{1pt}{1pt}
\textit{Music-to-Dance (30s long-term dance, Lodge)} & 45.56 & 34.29 & 6.75 & 0.114  & 46.772 & 29.857 \\
\textit{Music-to-Dance (30s long-term dance, Lodge-Morph)} & 43.96 & 32.88 & 6.90 & 0.083 & 0.000 & 3.163 \\
\hline
\hline
\specialrule{0em}{1pt}{1pt}
\multicolumn{7}{c}{PHC-based baseline} \\
\specialrule{0em}{1pt}{1pt}
\hline
\specialrule{0em}{1pt}{1pt}
\textit{Text-to-Motion (MoMask+PHC)} & 0.183 & - & 0.785 & 0.749 & 0.000 & 2.451  \\
\textit{Text-to-Motion (MoMask+Morph)} & \textbf{0.041} & - & \textbf{0.816} & \textbf{0.647 }& \textbf{0.000} & \textbf{2.141}  \\
\hline
\end{tabular}}

\end{table*}

\section{Semantic Alignment Analysis} \label{multi_round}
As shown in Fig.~\ref{fig:tsne}, Morph significantly outperforms MG on alignment metrics. Globally, Morph demonstrates superior realism by closely matching the statistical distribution of real motions—exhibiting similar clustering patterns and range of variation (Left). Crucially, it also achieves stronger semantic alignment with input text features, forming tighter clusters around corresponding text embeddings to better capture intended meanings (Middle). Locally, Morph provides enhanced semantic matching at the segment level, ensuring fine-grained motion elements correspond more accurately to detailed text semantics throughout the sequence (Right). In summary, Morph can generate semantically faithful, realistic motions compared to the MG baseline.

\section{Cross-Task Generalization Ability}
To evaluate the cross-task generalization of the MPR module, we conducted cross-validation by testing its performance across two distinct tasks: text-to-motion (using the MoMask dataset) and music-to-dance (using Bailando). These tasks differ significantly in input modalities—one driven by linguistic descriptions, the other by rhythmic audio-and in motion characteristics, from daily actions to stylized dance moves. As shown in Tab.~\ref{tab:crosstask}, the MPR module retains strong performance even when trained without task-specific synthetic data: it not only preserves motion quality (e.g., smooth transitions and natural postures) but also maintains physical plausibility (avoiding joint distortions or gravity-defying movements). This confirms its ability to generalize beyond specific task boundaries.

\section{User Study}
In the user study, we evaluated win rates across four critical dimensions—semantic alignment (matching text descriptions), authenticity (resembling real motions), physical plausibility (avoiding unnatural joint movements), and fluency (smooth temporal transitions). Morph decisively outperformed baseline methods here: as shown in Tab.~\ref{tab:win_rate}, it achieved significantly higher win rates of 87.4\%, 85.2\%, 96.2\%, and 82.4\% respectively. Such consistent leads across all key metrics confirm its comprehensive advantages in motion generation quality.

\section{More Qualitative Results} \label{qualitative_comparisons}
\begin{figure*}[!t]
	\centering
	\tabcolsep=0.05cm
    \includegraphics[width=0.9\linewidth]{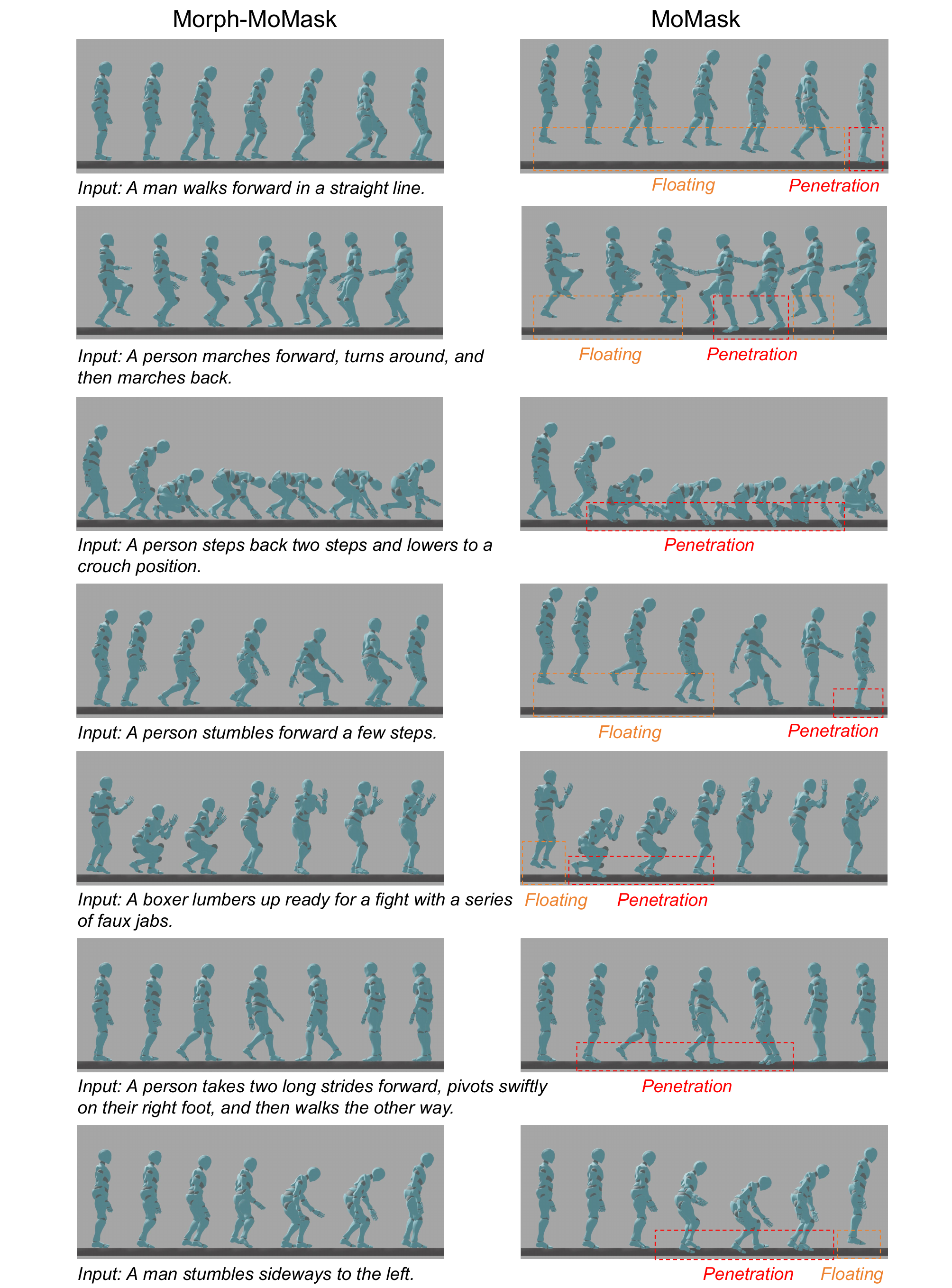}
    \caption{Qualitative comparisons for text-to-motion on HumanML3D test set between Morph-MoMask and MoMask.}
\label{fig:visualization_t2m_appendix_morph_momask}
\end{figure*}

\begin{figure*}[!t]
	\centering
	\tabcolsep=0.05cm
    \includegraphics[width=0.9\linewidth]{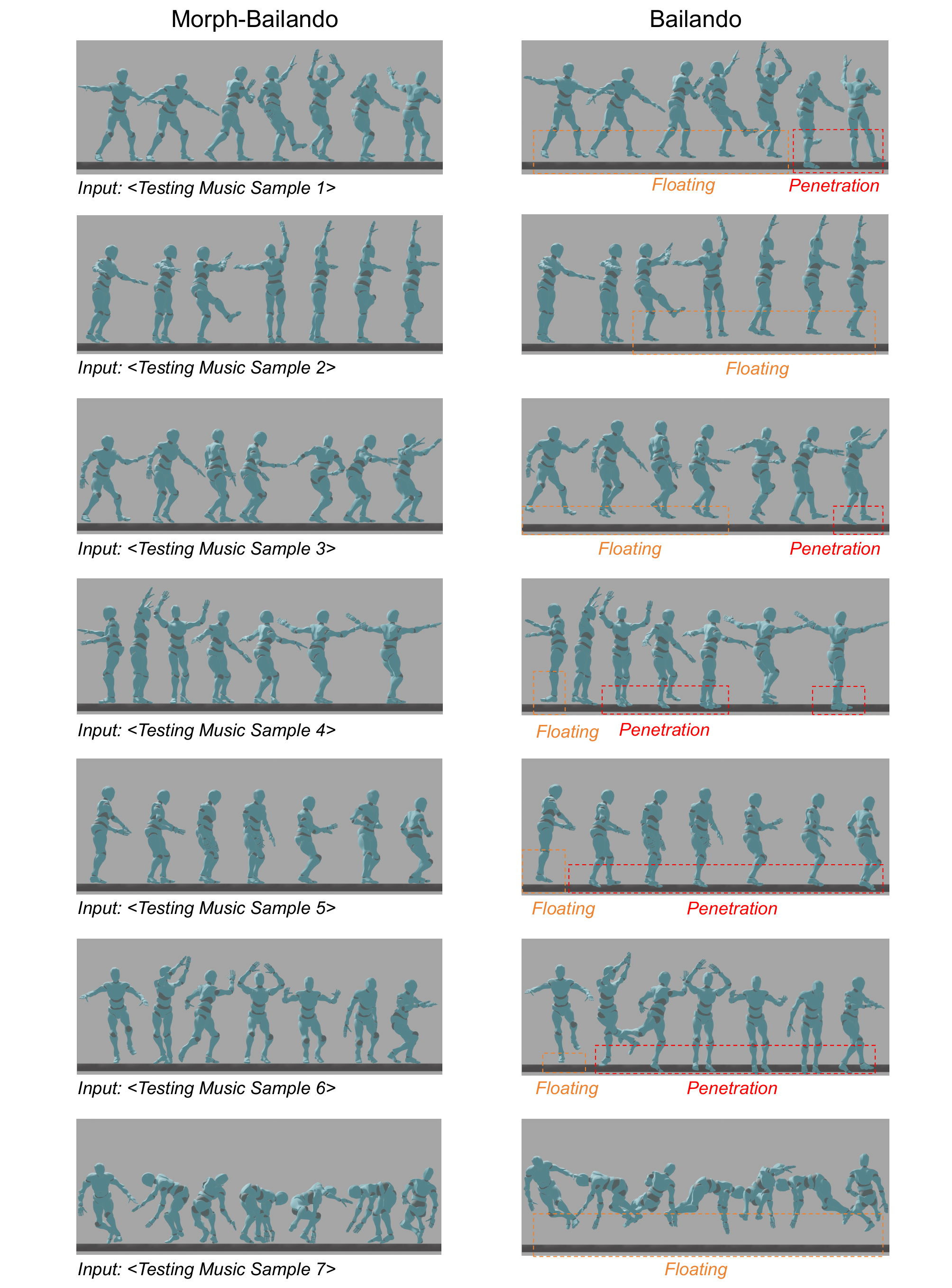}
    \caption{Qualitative comparisons for music-to-dance on AIST++ test set between Morph-Bailando and Bailando. For music-to-dance, the testing music samples will be used as inputs. }
\label{fig:visualization_a2d_appendix_morph_bailando}
\end{figure*}

Fig.~\ref{fig:visualization_t2m_appendix_morph_momask} and Fig.~\ref{fig:visualization_a2d_appendix_morph_bailando} provide the additional qualitative results for the text-to-motion and music-to-dance generation tasks using Morph.

As shown in Fig.~\ref{fig:visualization_t2m_appendix_morph_momask}, in the text-to-motion generation task, floating and penetration are common artifacts in motion generation, often resulting from inaccuracies in the estimation of translation. 
However, Morph effectively addresses these issues, successfully mimicking  the input motion and demonstrating  a significant improvement in mitigating these artifacts. The generated motions are both physically plausible and realistic, showcasing Morph's enhanced performance in this task.

As shown in Fig.~\ref{fig:visualization_a2d_appendix_morph_bailando}, in the music-to-dance generation task, 
floating and penetration are the most prominent issues. 
Due to the faster frequency of dance movements, these artifacts occur more frequently. Morph effectively mitigates these issues, generating motions that are not only physically plausible but also exhibit a higher degree of realism. 

In summary, Morph demonstrates significant improvements in both the text-to-motion and music-to-dance tasks. By accurately estimating  translational motion, Morph is able to generate motions that are not only physically feasible but also exhibit a higher degree of realism.

